\documentclass[twocolumn,10pt]{article}
\usepackage{icicpe}
\usepackage[utf8]{inputenc}
\usepackage[pdfencoding=auto]{hyperref}
\usepackage{url}
\usepackage{amsmath,amssymb}
\usepackage{graphicx}
\usepackage{booktabs}
\usepackage[symbol]{footmisc}

\DeclareMathOperator{\rank}{rank}

\title[english]{Spectral Stability of Pseudoinverse-Based Extreme Learning Machine}

\author[english]{
Nguyen Bich Van$^{1,\ast}$, Khong Ngoc Anh$^{2}$\\
$^1,^2$: Institute for Artificial Intelligence, VNU University of Engineering and Technology, Vietnam\\
\texttt{$^1$: nbvan@vnu.edu.vn, $^2$: 22022549@vnu.edu.vn}
}

\finalcopy

\begin{document}

\twocolumn[
\maketitle
\begin{abstract}
Extreme Learning Machine (ELM) computes output weights analytically using the Moore--Penrose pseudoinverse. Although this leads to fast training, its numerical stability depends strongly on the conditioning of the hidden-layer matrix. This paper studies pseudoinverse-based ELM from a spectral perspective. We show that the smallest singular value governs perturbation amplification in the output weights, while the condition number provides a quantitative measure of hidden-layer instability. We compare SVD-based pseudoinverse computation with iterative hyperpower methods and discuss width-dependent conditioning through a random feature interpretation. Experiments on synthetic matrices and ELM benchmarks show that SVD-based methods remain the most reliable under ill-conditioning, while iterative methods are more sensitive to spectral properties. The results suggest that ELM stability is fundamentally governed by the singular-value structure of the hidden-layer matrix.
\end{abstract}

]
\footnotetext[1]{Corresponding author}

\section{Introduction}

Extreme Learning Machine (ELM), originally introduced by Huang et al.~\cite{Huang}, is a single-hidden-layer feedforward neural network in which hidden-layer parameters are randomly assigned and output weights are computed analytically. Given a training set $\{(x_j,t_j)\}_{j=1}^N$, the hidden-layer matrix is
\begin{equation}
H_{ji}=g(w_i^T x_j+b_i),
\end{equation}
where $w_i$, $b_i$, and $g$ denote the hidden weights, biases, and activation function, respectively. The output weights are obtained from
\begin{equation}
H\beta=T, \qquad \beta=H^+T,
\end{equation}
where $H^+$ is the Moore--Penrose pseudoinverse.

This closed-form training is computationally attractive, but it is also sensitive to the spectral properties of $H$. If the hidden-layer matrix is ill-conditioned, small singular values may amplify perturbations in the targets, numerical roundoff, or solver approximation errors. Thus, the stability of ELM is not only a learning problem but also a numerical linear algebra problem.

Related work on ELM has emphasized fast analytical training and classification performance~\cite{Huang}. The theory of generalized inverses and least-squares solutions is classical~\cite{BenIsrael}, while SVD-based conditioning and stability analysis are standard topics in numerical linear algebra~\cite{Golub,Trefethen}. Iterative inverse and pseudoinverse methods, including Newton--Schulz and hyperpower iterations, may be computationally attractive but can be sensitive to initialization and ill-conditioning. In parallel, random matrix theory provides useful estimates for the singular values of random feature matrices~\cite{RV,Vershynin}.

The contribution of this paper is a compact spectral interpretation of pseudoinverse-based ELM stability. We connect:

$
\text{hidden-layer spectrum}\rightarrow \text{conditioning}\rightarrow
\text{pseudoinverse sensitivity}\rightarrow \text{learning stability}.
$

We also summarize experiments showing how singular values, condition numbers, and solver behavior are related in synthetic matrices and ELM benchmarks.

\section{Spectral Stability Analysis}

Let $H\in\mathbb{R}^{N\times L}$ be the hidden-layer matrix of an ELM and suppose its compact singular value decomposition is
\begin{equation}
H=U_r\Sigma_rV_r^T,
\end{equation}
where $r=\rank(H),$ $U_r\in \mathbb{R}^{N\times r}, V_r\in \mathbb{R}^{L\times r}$ have orthonormal columns, $\Sigma_r=\operatorname{diag}(\sigma_1,\ldots,\sigma_r)$ and $\sigma_{max}(H)\geq \sigma_1\geq\cdots\geq\sigma_r=\sigma_{min}(H)>0$ are positive singular values of $H$. The Moore--Penrose pseudoinverse is
\begin{equation}
H^+=V_r\Sigma_r^{-1}U_r^T.
\end{equation}
Hence,
\begin{equation}
\|H^+\|_2=\frac{1}{\sigma_{\min}(H)}.
\label{eq:pinv_norm}
\end{equation}
Equation~\eqref{eq:pinv_norm} shows that the smallest singular value is the key spectral quantity controlling pseudoinverse amplification.

The condition number of $H$ is
\begin{equation}
\kappa(H)=\frac{\sigma_{\max}(H)}{\sigma_{\min}(H)}.
\end{equation}
It measures how unevenly $H$ stretches different directions. If $\kappa(H)$ is large, at least one direction is nearly collapsed by $H$, making the least-squares solution sensitive to perturbations.

Consider a perturbation $T\mapsto T+\Delta T$. The corresponding perturbation in the ELM output weights is
\begin{equation}
\Delta\beta=H^+\Delta T.
\end{equation}
Therefore,
\begin{equation}
\|\Delta\beta\|_2\leq \|H^+\|_2\|\Delta T\|_2
=\frac{\|\Delta T\|_2}{\sigma_{\min}(H)}.
\label{eq:perturbation}
\end{equation}
Thus, small $\sigma_{\min}(H)$ or large $\kappa(H)$ implies stronger perturbation amplification. This provides the main theoretical explanation for instability in pseudoinverse-based ELM.

\section{Solvers and Width-Dependent Conditioning}

SVD-based pseudoinverse computation directly uses the singular values of $H$ and is typically backward stable because it relies on orthogonal transformations~\cite{Golub,Trefethen} which preserve Euclidean norm. It also allows small singular directions to be detected and truncated if necessary.

Iterative methods approximate the pseudoinverse through repeated matrix updates. A representative example is the Newton--Schulz iteration
\begin{equation}
X_{k+1}=X_k(2I-HX_k).
\end{equation}
Such methods can be efficient in favorable settings, but their convergence depends on the initialization, scaling, stopping criteria, and the spectrum of $H$. When $H$ is ill-conditioned, convergence may slow down, stagnate, or become sensitive to finite-precision errors.

The hidden-layer matrix in ELM may be viewed as a random feature matrix because
\begin{equation}
H_{ji}=g(w_i^Tx_j+b_i)
\end{equation}
is generated from random hidden parameters such as hidden weights $w_i$ and bias $b_i$. If the normalized entries of H are independent, centered, sub-Gaussian random variables with unit
variance, following a standard theorem in random matrix theory \cite{RV,Vershynin} we have
\begin{equation}
\sigma_{\min}(H)\geq c(\sqrt{N}-\sqrt{L})
\label{eq:width_law}
\end{equation}
with probability at least $1-2\exp(-C(N-L))$ for some constants $c,C>0$. Consequently, $\|H^+\|_2\leq \frac{1}{c(\sqrt{N}-\sqrt{L})}.$

Although practical ELM matrices are not fully independent random matrices, this estimate motivates a width-dependent interpretation: larger sample size tends to improve stability, while excessively large hidden width may deteriorate conditioning.

\section{Experimental Results}

We conduct two groups of experiments. The first group uses synthetic matrices with prescribed singular-value spectra to compare SVD-based and iterative pseudoinverse solvers. The second group evaluates pseudoinverse-based ELM on benchmark classification datasets, including MNIST, Fashion-MNIST, and ISOLET.

The main evaluation metrics are least-squares residual, Penrose residuals, runtime, convergence success, classification accuracy, the smallest singular value $\sigma_{\min}(H)$, and the condition number $\kappa(H)$.

Table~\ref{tab:iterative_convergence} summarizes the convergence behavior of iterative pseudoinverse methods under different spectral conditions. The results show that iterative methods are reliable for well-conditioned and moderately ill-conditioned spectra, but fail completely for severely ill-conditioned spectra. In particular, all iterative runs succeeded in the well-conditioned case, whereas no iterative method succeeded in the severely ill-conditioned case. This confirms that convergence reliability is strongly controlled by the singular-value structure of the matrix.

\begin{table}[t]
\centering
\caption{Convergence behavior of iterative pseudoinverse methods under different spectral conditions. $10^{-10}$ and $10^{-7}$ are thresholds.}
\label{tab:iterative_convergence}
\begin{tabular}{lcc}
\toprule
Spectrum & $10^{-10}$  & $10^{-7}$ \\
\midrule
Well-conditioned & 36/36 & 36/36 \\
Moderately ill-conditioned & 36/36 & 36/36 \\
Severely ill-conditioned & 0/36 & 0/36 \\
Rank deficient & 10/36 & 36/36 \\
Clustered small & 24/36 & 36/36 \\
\bottomrule
\end{tabular}
\end{table}

The detailed residual summaries further support this observation. In the well-conditioned case, all methods achieved successful runs with residuals close to machine precision. For example, Newton--Schulz achieved a median least-squares residual of $1.21\times 10^{-15}$ under $\varepsilon_{\mathrm{conv}}=10^{-10}$. In contrast, under the severely ill-conditioned spectrum, all iterative methods failed, while SVD-based methods retained successful runs.

\begin{figure}[t]
\centering
\includegraphics[width=0.46\textwidth]{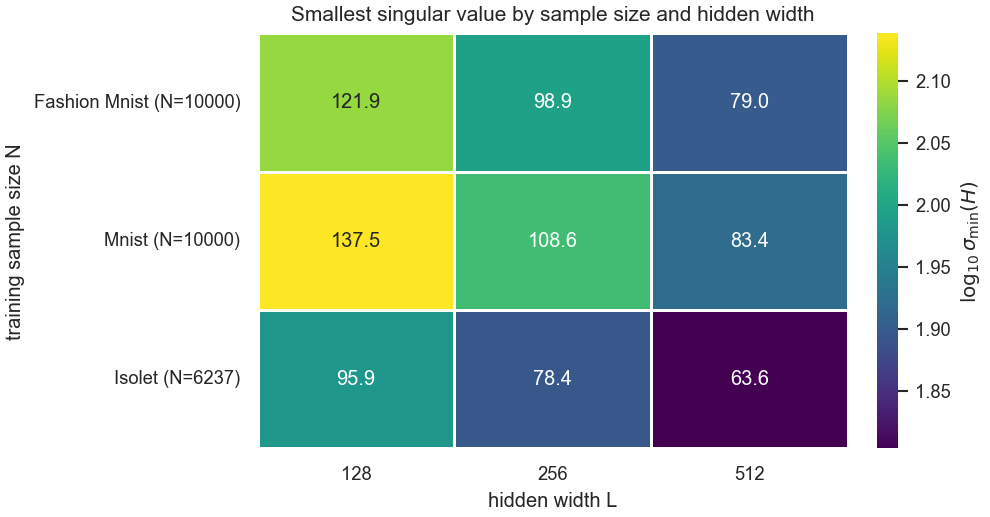}
\caption{Mean smallest singular value of the ELM hidden-layer matrix for different training sample sizes \(N\) and hidden widths \(L\). Larger \(L\) generally reduces \(\sigma_{\min}(H)\), indicating poorer conditioning and stronger pseudoinverse sensitivity.}
\label{fig:heatmap_sigmamin}
\end{figure}

Figure~\ref{fig:heatmap_sigmamin} reports the mean smallest singular value of the ELM hidden-layer matrix under different sample sizes and hidden widths. The results show a consistent decrease of \(\sigma_{\min}(H)\) as the hidden width \(L\) increases. This behavior supports the width-dependent spectral interpretation: increasing \(L\) may improve representational capacity, but it can also worsen conditioning and increase the sensitivity of the pseudoinverse solution.

\section{Conclusion and Future Work}

This paper studied pseudoinverse-based ELM from a spectral stability perspective. The analysis showed that the smallest singular value determines perturbation amplification in the output weights, while the condition number provides a compact measure of hidden-layer ill-conditioning. The comparison between SVD-based and iterative methods indicates that SVD remains the most reliable solver under severe ill-conditioning, whereas hyperpower methods are more sensitive to spectral properties and initialization.

The width-dependent interpretation further suggests that hidden-layer stability depends on the balance between sample size and hidden width. Future work includes regularized ELM, randomized SVD, improved stopping rules for iterative solvers, and larger-scale GPU implementations.

\end{document}